\pdfoutput=1  
\documentclass[conference]{IEEEtran}
\IEEEoverridecommandlockouts

\usepackage{cite}
\usepackage{amsmath,amssymb,amsfonts}
\usepackage{amsthm}
\usepackage{graphicx}
\usepackage{textcomp}
\usepackage{xcolor}
\usepackage{multirow}
\usepackage{booktabs}
\usepackage{array}
\usepackage{url}
\usepackage{soul}
\usepackage{xstring}
\usepackage{dsfont}
\usepackage{enumitem}
\usepackage{pifont}
\usepackage[caption=false,font=footnotesize]{subfig}
\usepackage[export]{adjustbox}

\setul{1pt}{.4pt}
\newcommand{\PP}[1]{\smallskip\noindent{\bf \IfEndWith{#1}{.}{#1}{#1.}}}

\newcommand{\todo}[1]{}
\newcommand{\TODO}[1]{}

\let\textcite\cite

\begin{document}

\title{ARMOR: Manifold-Oriented Training for\\ Adversarially Robust Aerial Object Detection under Data Scarcity%
\thanks{\textsuperscript{*}Equal contribution.}}

\author{%
\IEEEauthorblockN{Haoran Wang\textsuperscript{1,*}, Matthew Lau\textsuperscript{1,*}, Alec Helbling\textsuperscript{1}, Matthew Hull\textsuperscript{1}, ShengYun Peng\textsuperscript{1}, Mansi Phute\textsuperscript{1},\\
Martin Andreoni\textsuperscript{2}, Willian T. Lunardi\textsuperscript{3}, Duen Horng Chau\textsuperscript{1}, Wenke Lee\textsuperscript{1}}
\IEEEauthorblockA{\textsuperscript{1}\textit{Georgia Institute of Technology}, Atlanta, GA, USA \qquad
\textsuperscript{2}\textit{Khalifa University}, Abu Dhabi, UAE\\
\textsuperscript{3}\textit{Technology Innovation Institute}, Abu Dhabi, UAE\\
\{haoran.wang, mlau40, alechelbling, matthewhull, speng65, mphute6, polo\}@gatech.edu, wenke@cc.gatech.edu\\
martin.andreoni@ku.ac.ae, williantessarolunardi@gmail.com}
}

\maketitle

\begin{abstract}
Aerial object detection is increasingly deployed in real-world applications, but models remain vulnerable to physical, universal adversarial patches that cause them to miss objects.
Furthermore, defenders face the practical constraint of training data scarcity: aerial imagery is costly to collect and label, so a deployment site typically yields hundreds of images rather than the tens of thousands that adversarial robustness benchmarks assume.
To tackle model vulnerability \textit{and} training data scarcity, we propose Adversarial Robustness with Manifold-Oriented Training (ARMOR), a novel defense that realizes the core insights of on-manifold adversarial training (OMAT) in low-data regimes.
ARMOR builds on the insight of OMAT to model the data manifold --- the compact structure capturing the data's relevant features --- to learn and robustify these features during training.
While OMAT relies on the data-intensive operations of training large generative models and adversarial training to achieve this, ARMOR adopts a data-efficient approach that reuses labels the detection task already supplies:
ARMOR (i) masks image backgrounds to retain object-relevant features, and (ii) injects randomized patches on objects to improve feature robustness.
Our low-data experiments with physically-realizable adversarial patches evaluate both query-free transfer attacks and defense-aware attacks.
ARMOR maintains strong clean performance of over 0.90 model confidence, while improving adversarial robustness by up to 0.32 in model confidence over state-of-the-art defenses.
Physical experiments with printed patches confirm that these gains survive deployment.
Overall, ARMOR translates insights from manifold-based training to defend object detectors amidst training data scarcity.
\end{abstract}


\section{Introduction}
\label{section:AdvML_Intro}

Object detection models are progressively deployed in real-world applications of aerial imagery, such as traffic monitoring \cite{physical_aerial_adv_att}, environmental protection \cite{uav_obj_det_survey} and disaster relief \cite{drone_object_det}.
Drones and fixed aerial cameras stream imagery continuously, and the detector sits at the ingest layer of the pipeline: every downstream count, track and alert is derived from what it does or does not detect.
Despite their growing adoption, these models remain vulnerable: failures in perception systems have contributed to accidents \cite{ai_failure_auto_vehicle}.
More disconcertingly, these vulnerabilities can be straightforwardly exploited in aerial imagery with physical, universal adversarial attack patches that fool object detectors \cite{physical_aerial_adv_att}.
A single printed patch is therefore a cheap and persistent way to corrupt the pipeline at its source --- a \textit{veracity} failure that no amount of downstream data volume can recover.
Besides the ease of attacks, the attacker can also generate model-agnostic ``grey-box'' attacks \cite{explaining_harnessing_adv_ex_goodfellow_iclr2015,why_adv_att_transfer_usenix2019}, which require far less knowledge than a defense-aware ``white-box'' threat model.
Despite these vulnerabilities, the viability of defenses in aerial imagery has been understudied \cite{survery_adv_att_def_autonomous_veh_app}.
As such, improving the adversarial robustness of aerial object detectors is vital.

\begin{figure}[t]
    \centering
    \includegraphics[clip, width=\columnwidth]{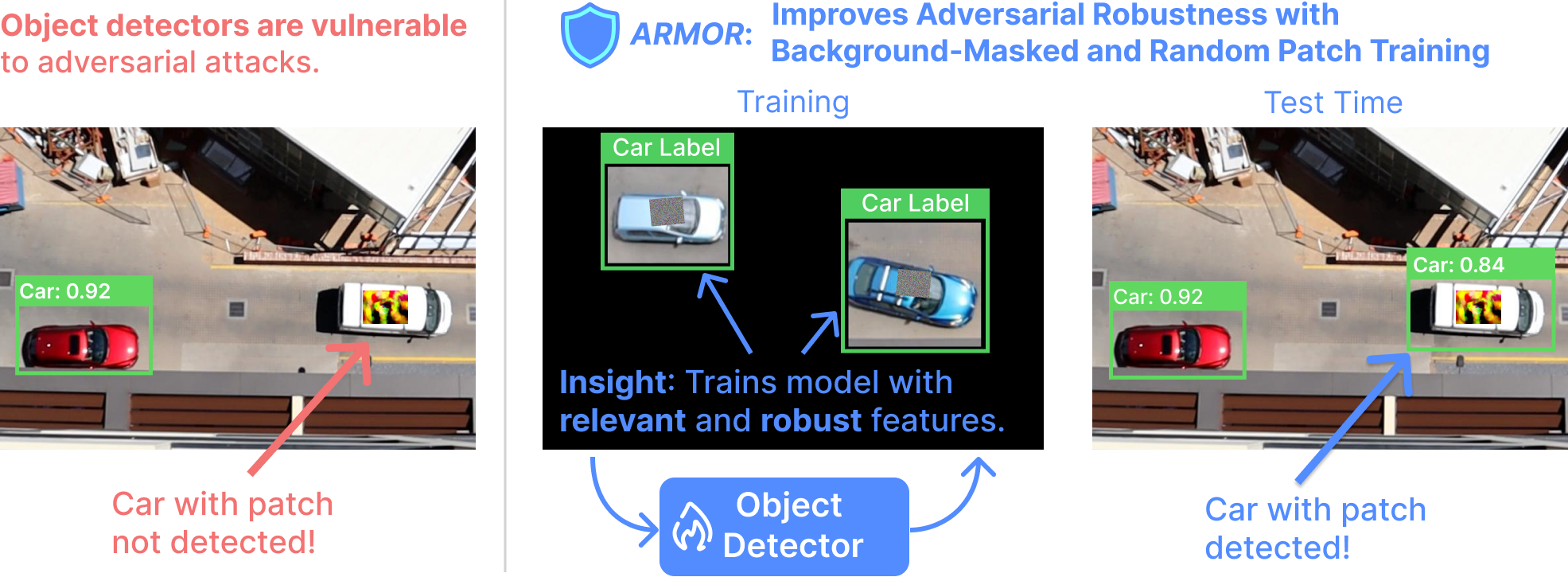}
    \caption{\textbf{Main idea of ARMOR.}
    We model the two data-intensive operations of on-manifold adversarial training --- on-manifold projection and adversarial training --- with background removal and random patches respectively.
    These modifications force the model to \textit{learn relevant and robust features}, defending physically-realizable adversarial patch attacks.
    }
    \label{fig:crown_jewel_main}
\end{figure}

Moreover, robustness alone is not enough.
In many aerial imagery applications, defenders face a second, underexplored challenge: \textbf{training or fine-tuning data is often scarce} due to high costs and limited availability \cite{label_aerial_imgs_icml2012, aerial_limited_training_data_sat_img_ml}.
The scarcity is one of \textit{labels}, not pixels: raw aerial footage is abundant, but each deployment site has its own camera geometry, altitude and lighting, so the labeled set matching the site must be annotated by hand and stays small.
Many existing defenses have yet to account for this, because benchmarks offer abundant data.
To illustrate, CIFAR-10, a popular benchmark for studying adversarial robustness \cite{adv_att_obfuscated_grad, robust_overfitting_OG_kolter}, offers 60,000 images for classification at $32\times32$ dimensionality \cite{cifar10}.
In contrast, aerial object detection datasets often contain only hundreds to tens of thousands of images \cite{physical_aerial_adv_att}, despite much higher dimensionality of $320\times320$ to $640\times640$ \cite{yolov3, yolo11_ultralytics}.
It is not clear if defenses can generalize to data-scarce settings.
Thus, we address the practical question: \textbf{How can we improve the adversarial robustness, while maintaining good clean performance, of aerial object detection when training data is scarce?}

One promising approach to improve adversarial robustness is on-manifold adversarial training (OMAT) \cite{disentangle_adv_robustness_generalisation, dual_manifold_adv_robustness_lin2020}.
OMAT aims to sidestep the typical trade-off between adversarial robustness and generalization (i.e., clean performance) --- known as robust overfitting \cite{robust_overfitting_OG_kolter} --- by leveraging the manifold hypothesis.
The manifold hypothesis is a common assumption in deep learning, positing that real-world, high-dimensional data like images lie on a low dimensional structure or ``manifold'' \cite{deep_learning_manifold_disentanglement}.
Intuitively, this manifold captures the relevant features in realistic data.
However, OMAT is impractical in low-data regimes due to its reliance on training large generative models to approximate the manifold \cite{training_GANs_limited_data} and solving expensive adversarial optimization problems for robustness \cite{goodfellow_neurips2016_tutorial_GANs, alg_stability_adv_training_neurips_xing2021, few_shot_adv_robustness_meta_learning_neurips2020}.

To overcome this limitation, we propose \textbf{ARMOR} (\ul{A}dversarial \ul{R}obustness with \ul{M}anifold-\ul{Or}iented Training, Figure \ref{fig:crown_jewel_main}), a defense that retains OMAT's benefits in real-world, data-scarce applications.
Instead of OMAT's data-intensive operations of generative modeling and adversarial training, ARMOR is based on OMAT's insight on guiding models to learn \textit{relevant} and \textit{robust} features.
Crucially, ARMOR sources both operations from annotation the detection task already pays for --- bounding box labels --- so the defense adds \textit{no labeling cost} over standard fine-tuning.
To \textit{learn relevant features}, instead of using generative models, ARMOR masks image backgrounds defined by bounding box labels to a uniform black during training.
Irrelevant background features are removed, focusing the model on object-relevant features.
To \textit{learn robust features}, instead of adversarial training, ARMOR inserts randomized patches on objects, approximated by the center of each bounding box, during training.
Randomization simulates variability due to noise, improving feature robustness without the instability of min-max adversarial optimization.

These two steps challenge the conventional wisdom that distribution shift between train and test data degrades performance: ARMOR deliberately introduces one, since test images contain backgrounds and have no patches on objects (Figure \ref{fig:crown_jewel_main}).
Nonetheless, ARMOR consistently increases robustness to adversarial attacks while retaining strong accuracy on clean data, and we read this mitigation of robust overfitting as evidence that ARMOR is a data-efficient adaptation of OMAT's insights.
Our contributions are as follows:
\begin{itemize}[leftmargin=*, noitemsep, topsep=2pt]
    \item We propose ARMOR, a novel defense for aerial object detectors against physically-realizable adversarial patches that accounts for training data scarcity.
    \item To mitigate training data scarcity, ARMOR adapts the two data-intensive operations of on-manifold adversarial training.
    Instead of (i) generative modeling and (ii) adversarial training, ARMOR uses bounding box labels to (i) mask image backgrounds and (ii) insert random patches on objects during training.
    These operations guide the model to learn relevant and robust features in a data-efficient manner.
    \item Despite training data scarcity, our experiments demonstrate that ARMOR is an effective defense, maintaining strong clean performance of over 0.90 model confidence while improving adversarial robustness by up to 0.32 in model confidence over state-of-the-art defenses.
    A manifold analysis of the learned representations explains \textit{why}, and physical experiments with printed patches confirm the gains hold in deployment.\looseness=-1
\end{itemize}

\section{Related Works}

\PP{Object Detection}
The goal of object detection is to localize objects of interest in an image by enclosing the object with a bounding box.
We focus on the You Only Look Once (YOLO) v3 object detector \cite{yolov3} because of its success in aerial imagery \cite{aerial_obj_det_robustness}, and additionally evaluate the recent YOLOv11 \cite{yolo11_ultralytics}.
Aerial images are obtained from a bird's-eye view camera, usually mounted on a drone or satellite, enabling observation of large areas for traffic and environmental monitoring \cite{physical_aerial_adv_att, uav_obj_det_survey} and self-driving \cite{auto_driving_aerial_imagery_iros2024}.
\label{section:aerial_imagery_lit_review}
However, aerial object detection is less researched than ground-based detection --- common detectors like YOLO are pre-trained on ground-based datasets \cite{YOLO_OG_Redmon_2016_CVPR, yolov3}, and objects may be low resolution due to small size or weather \cite{aerial_obj_det_robustness}, challenging both clean transfer and adversarial robustness \cite{adv_att_aerial_agg_det}.\looseness=-1

\PP{Adversarial Patch Attacks}
Adversarial attacks make perturbations on images to change the model's prediction.
We consider untargeted attacks \cite{disentangle_adv_robustness_generalisation, ShapeShifter_Chen2019}, which change images to maximize the loss on the altered images.
Formally, let $f:\mathcal X \to \mathbb R^m$ be the model, $\mathcal Y$ be the label space, $(\mathbf x, y)\in \mathcal X \times \mathcal Y$ be the ground-truth datum-label pair and $\mathcal L:\mathbb R^m \times \mathcal Y \to \mathbb R_{\geq 0}$ be the loss function.
The adversarial perturbation $\delta$ solves $\max_\delta \mathcal{L}(f(\mathbf x + \delta), y)$ subject to $\delta \in \Delta$, where $\Delta$ is the constraint set on $\delta$.
Adversarial patches \cite{brown2018adversarialpatch} restrict $\Delta$ to localized regions, providing a realistic threat model: attackers print physical patches to place in the environment.
We study this in aerial imagery because (i) the attack is practical --- attackers can apply patches on their own car at a safe distance, and (ii) aerial patches are less susceptible to occlusions or camera angle changes, enabling straightforward transfer from digital to physical attacks.
Physically-realizable patch generation against object detectors \cite{phys_adv_ex_object_detectors_usenix} is generally harder than against classifiers \cite{ShapeShifter_Chen2019}.
Expectation over Transformation (EoT) \cite{EOT_ICML_pmlr-v80-athalye18b} is commonly applied for robustness against physical transformations \cite{ShapeShifter_Chen2019, physical_aerial_adv_att}.
\textcite{physical_aerial_adv_att} attacks YOLOv3 by minimizing model confidence with total variation (TV) and non-printability score (NPS) regularization.

\subsection{Adversarial Defenses}\label{section:adv_ML_def}

\PP{Pre-processing}\label{section:pre-processing_image}
Some defenses localize and mask adversarial patches.
SAC \cite{SAC_Liu_2022_CVPR} trains a model to detect patches, but \textit{new patches} can be developed against them \cite{PAD_Jing_2024_CVPR}, and more fundamentally \textit{deep learning detectors can be easily evaded} \cite{carlini_wagner_detection}.
Information-theoretic approaches avoid neural network vulnerabilities: Jedi \cite{Jedi_Tarchoun_2023_CVPR} uses entropy but requires threshold estimation, while PAD \cite{PAD_Jing_2024_CVPR} outperforms SAC, Jedi and ObjectSeeker using mutual information and compression quality, optionally refined with SAM \cite{segment_anything_model}, but relies on the segmentation model to segment patches it may not be trained on.
Traditional image-processing defenses such as JPEG compression \cite{SHIELD_JPEG_Defense, guo2018_adv_def_input_transforms} and image quilting \cite{guo2018_adv_def_input_transforms} suffer from \textit{robust overfitting} and have been broken by defense-aware attacks \cite{adv_att_obfuscated_grad}; physical patches are also designed to be smooth enough to survive compression.
Overall, these techniques are useful but \textit{do not directly} address model vulnerability.

\PP{Model Robustness}\label{section:model_robustness}
Data augmentations encode invariances that improve robustness, but adversarial performance remains 20\% lower than clean \cite{data_augmentation_adv_robustness_neurips2021}, making them \textit{insufficient} on their own.
RobustDet \cite{RobustDet_Dong-ECCV_2022} modifies adversarial training for object detection but requires training from scratch with extensive data, and \textcite{adv_training_obj_det} suffers a 36\% decrease in clean mAP.
\textcite{Shafahi_ICLR_2020_Adversarially_Robust_Transfer_Learning} focuses on low-data fine-tuning but requires a robust pre-trained model, which may be \textit{unavailable} in practice, especially for object detection.

\PP{Distinguishing Off- and On-Manifold Attacks}\label{section:off_and_on_manifold_attack_defense}
There is a theory that dichotomizes adversarial examples into those on and off the data manifold --- the collection of relevant features in realistic data.
Off-manifold vulnerability arises from perturbations targeting \textit{irrelevant} features like background \cite{limitations_cond_generative_models, nonrobust_features_not_useful_one_class}.
Direct adversarial training in pixel space \cite{adv_training_madry2018} is popular \cite{alg_stability_adv_training_neurips_xing2021, adv_att_obfuscated_grad}, but only alleviates off-manifold vulnerability \cite{dual_manifold_adv_robustness_lin2020}: it defends only ``unrealistic'' background perturbations and suffers from robust overfitting \cite{robust_overfitting_OG_kolter}, with clean accuracy dropping by up to 20\%.
In contrast, on-manifold vulnerability arises from \textit{failure} to correctly use \textit{relevant} features \cite{disentangle_adv_robustness_generalisation, robust_eval_gen_models_BuzhinskyML2021}.
We study physically-realizable patches on objects, resembling on-manifold attacks (Section \ref{section:adv_att_threat_model}).
On-manifold adversarial training (OMAT) promotes robustness and generalization on toy problems \cite{disentangle_adv_robustness_generalisation, robust_eval_gen_models_BuzhinskyML2021} by projecting data onto the manifold then adversarial training.
However, both steps are \textit{difficult in low-data regimes}.
Manifold projection is typically approximated via generative models \cite{disentangle_adv_robustness_generalisation, dual_manifold_adv_robustness_lin2020}, but training requires $\sim$10,000 images even with transfer learning \cite{training_GANs_limited_data}, and off-the-shelf models lack guarantees for specific applications.
Adversarial training is also unstable \cite{goodfellow_neurips2016_tutorial_GANs, alg_stability_adv_training_neurips_xing2021}, exacerbated by low data \cite{few_shot_adv_robustness_meta_learning_neurips2020}.
\textit{This gap is exactly what ARMOR targets.}

\PP{Certified Robustness and Model Modification}\label{section:certified_robustness}
Certifiably robust defenses are rare in object detection: ObjectSeeker \cite{ObjectSeeker_SP_2023} breaks with multiple or large patches \cite{PAD_Jing_2024_CVPR}, and randomized smoothing \cite{randomized_smoothing_obj_det_neurips2020} aggregates predictions from noisy images but is \textit{impractical} due to thousands of inferences per frame, and only targets $\ell_p$ perturbations.
Separately, some defenses modify the detector itself \cite{Feature_Norm_Clipping_Yu_2021_ICCV, spatial_context_adv_robustness_obj_det_cvprw}, for instance by clipping the feature norms of intermediate layers \cite{Feature_Norm_Clipping_Yu_2021_ICCV}.
\label{section:model_modification}
These modifications are \textit{model-specific fixes} and, as \textcite{PAD_Jing_2024_CVPR} suggests, do not provide insights on defending object detectors in general.

\section{Problem Set-Up}

Our case study focuses on aerial imagery for car detection.
We study this setting due to its prevalence in applications such as traffic monitoring and self-driving \cite{physical_aerial_adv_att, auto_driving_aerial_imagery_iros2024}, its common feature of limited training data (Section \ref{section:AdvML_Intro}) and realistic threat model (Section \ref{section:adv_att_threat_model}).
Our main experiments use a stationary aerial camera that looks out onto the street for traffic monitoring.
We proceed to outline the attacker's and defender's goals.\looseness=-1

\begin{figure}[t]
    \centering
    \subfloat[ON patch on car roof.]{\includegraphics[height=1.35cm]{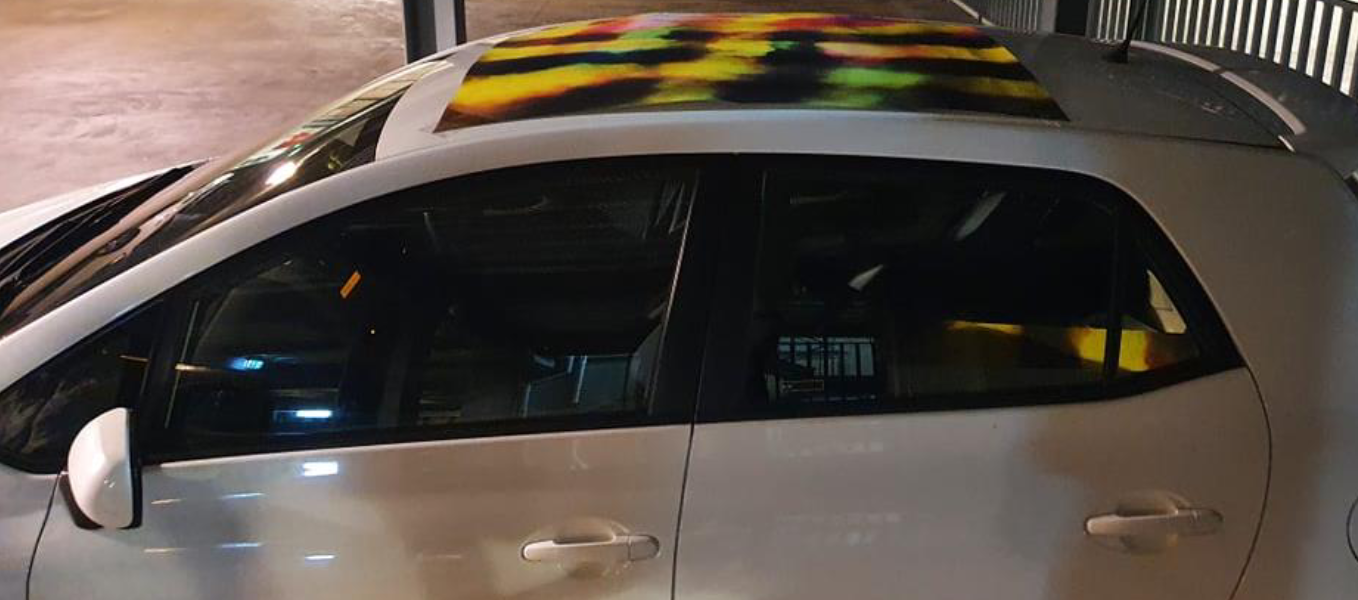}}
    \hfil
    \subfloat[Aerial view.]{\includegraphics[height=1.35cm]{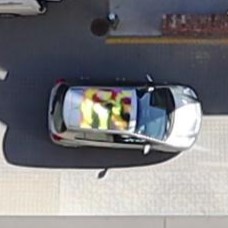}}
    \hfil
    \subfloat[ON-T patch \cite{physical_aerial_adv_att}.]{\includegraphics[height=1.35cm]{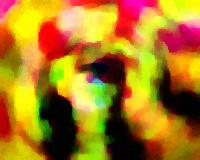}\label{fig:adv_patch_ON_sidestreet_gc}}
    \hfil
    \subfloat[ON-DA patch.]{\includegraphics[height=1.35cm]{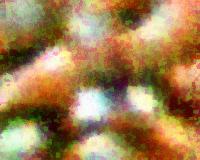}\label{fig:adv_patch_ON_sidestreet_gc_defense_aware}}
    \caption{Adversarial ``ON patches'' \cite{physical_aerial_adv_att} are easy for an attacker to deploy physically (a, b).
    Transfer (ON-T) and defense-aware (ON-DA) patches (c, d) are optimized for smoothness and printability.}
    \label{fig:adv_patches}
\end{figure}

\subsection{Threat Model: Realistic Evasion Attacks}
\label{section:adv_att_threat_model}

\PP{Attacker's Goal}
We follow the set-up from \textcite{physical_aerial_adv_att} on physically-realizable evasion attacks on aerial object detectors.
The attacker aims to cause their own car to evade detection of the object detector.
The attacker does so by placing an adversarial patch on the roof of the car, known as ``ON patches'' (Figure \ref{fig:adv_patches}).
The patch should also be robust to varying environmental conditions.
For the sake of evaluating attack efficacy, we do not consider partially hidden patches or cars.

\PP{Attacker's Knowledge}
We consider two levels of attack knowledge.
The first is the ``transfer'' attack (ON-T), traditionally known as grey-box.
Here, the attacker has access to a pre-trained model, which is often open-sourced, and can collect their own training data which, for simplicity, we assume to be the same as the defender's.
The attacker does not have knowledge about the defense strategies, and hopes to generate an adversarial patch that is model- and defense-agnostic \cite{explaining_harnessing_adv_ex_goodfellow_iclr2015,why_adv_att_transfer_usenix2019}.
Since this setting is the most practical for the attacker, we use transfer attacks to evaluate the \textit{practical feasibility of defenses}.
The second is the ``defense-aware'' attack (ON-DA), traditionally known as white-box, where the attacker has access to the defense, such as the fine-tuned model parameters and defense approach.
This attack stress-tests defenses, optimizing against the defense directly.
Since this setting is the \textit{most challenging} for the defender, we use defense-aware attacks to evaluate the robustness of defenses.

\begin{figure*}[t]
\centering
\includegraphics[width=.76\linewidth]{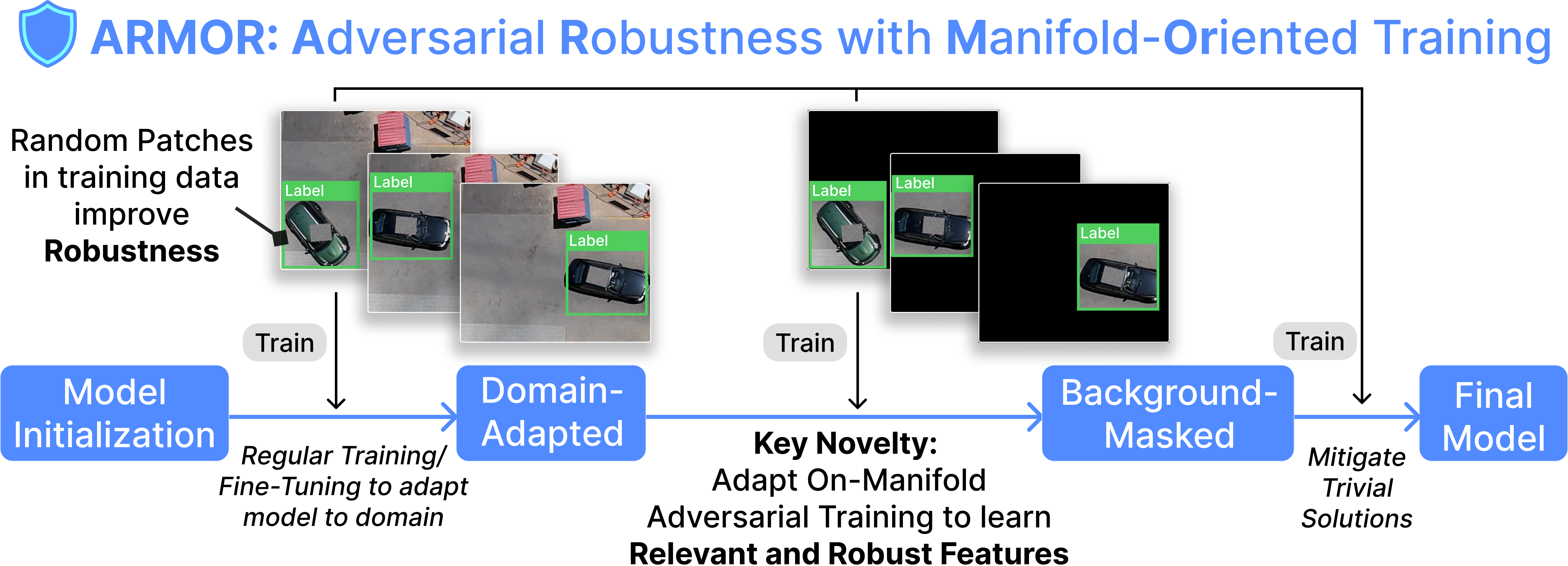}
\caption{\textbf{ARMOR's Method Diagram.}
ARMOR injects random patches onto objects during training to promote robustness.
The first phase of ARMOR's training is standard fine-tuning, which is used as a warm start for the second phase.
This second phase implements our key novelty of training with background-masked images and random patches.
The last phase of hybrid training mitigates any trivial solutions that the prior phase learns, such as localizing non-black regions as objects.
}
\label{fig:defense_summary}
\end{figure*}

\PP{Attacker's Capabilities}
Following \textcite{physical_aerial_adv_att}, the attacker optimizes patches digitally on training data, then prints and affixes the patch to the car roof.
The attacker cannot perturb the entire scene, but is restricted to modifying the patches.
We avoid the traditional terminology of ``grey-box'' and ``white-box'' because this \textit{physically-realizable attack} is weaker than the traditional setting, where two key assumptions no longer hold: the attacker cannot manipulate the whole image, only a small localized patch constrained to be on the object, and cannot ``cheat'' by using test-time data to generate patches.

\PP{Attacker's Strategy}
Following \textcite{physical_aerial_adv_att}, we allow \textit{only one patch} (universal attack) trained with \textit{physical deployment considerations}.
First, we ensure printability with a non-printability score (NPS) loss.
Second, we increase smoothness with a total variation (TV) loss to ensure that adversarial effects are not blurred out due to low resolution.
Third, to increase robustness to environmental variations, we generate attacks with EoT on the geometry (size and angle) and color (brightness, contrast and noise), since the patch can be in bright and dark spots (Figure \ref{fig:adv_patches}b).
To avoid obfuscated gradients, the attacker is also allowed to use BPDA by removing the output sigmoid activation \cite{adv_att_obfuscated_grad}.
Although adversarial robustness requires much more data than clean generalization, these constraints let us make more practical empirical observations: realistic threat models constrained to physically-realizable ON patches resemble on-manifold adversarial threats, not arbitrary threats such as pixel-level perturbations.

\subsection{Defender Model}
Like \textcite{physical_aerial_adv_att}, we assume that the defender has access to a model pre-trained on the task (car detection) and can collect some limited number of labeled training examples from the deployment site for fine-tuning, also known as transductive transfer learning.
To emphasize, datasets in this work have fewer than 1,000 training samples, while the popular benchmark CIFAR-10 \cite{cifar10} has 60,000 training samples.
The defender's budget is therefore an \textit{annotation} budget, and any defense that demands additional labeled data --- or a separately trained generative model --- is out of reach at the deployment site.

\section{Proposed Method: ARMOR}

The key idea of \ul{A}dversarial \ul{R}obustness with \ul{M}anifold-\ul{Or}iented Training (ARMOR) is shown in Figure \ref{fig:crown_jewel_main}, while a summary of the specific methodology is provided in Figure \ref{fig:defense_summary}.
ARMOR creates and uses novel adaptations of the two fundamental steps of OMAT for low-data applications.
First, for learning the projection onto the manifold, we remove the background of images with ground truth bounding box labels instead of training a generative model.
Second, to improve feature robustness, we place patches with randomly generated pixels on objects at every fine-tuning epoch rather than adversarially generate those pixels.
During testing, we feed raw (i.e., unmasked) images into the fine-tuned object detector.
Both steps reuse labels the detection task already provides, so ARMOR's data cost is that of ordinary fine-tuning.\looseness=-1

\subsection{Manifold Projection by Background Removal}
\label{section:on_manifold_proj_bg_rem}

The first step of OMAT is to project data onto the data manifold.
To model this step, we propose removing image backgrounds by masking pixels outside of bounding boxes to be black.
Background removal in image space during training is part of our novel contribution, and we justify it along the two axes of \textit{projection} and \textit{manifold}.

\PP{Projection}
Data augmentation applies multiple transformations to introduce a set of invariances the modeler wishes to capture, and increases dataset size as a side-effect (Section \ref{section:model_robustness}).
Background removal, in contrast, applies a single transformation and does not produce more data.
In fact, applying it a second time to the background-masked dataset returns the same dataset, which is reminiscent of the idempotency property of mathematical projections \cite{idempotent_projection_book}.
To be precise, background removal is an idempotent pixel-masking (off-object pixels set to a constant), not a linear projection; this idempotency is what makes it a practical approximation of an on-manifold projection.

\PP{Manifold}
After projection by OMAT, most information is lost and what remains are the relevant features captured by realistic data, with the split determined by the task.
In our case, the task is localizing objects of interest, so pixels pertaining to the object should be retained while everything else is information we can afford to lose.
This is also supported by how perturbations on the manifold only change object-relevant pixels in the classification literature \cite{disentangle_adv_robustness_generalisation}.
We may not have ground-truth masks of which pixels are important, but the object detection task supplies bounding box labels on the object's location.
We therefore use these labels to identify pixels in each bounding box as meaningful information while masking out other pixels to black --- in general, these pixels can be masked to anything, as long as they do not contain irrelevant background information.
Though the background may contain some context, this information can be spurious and many irrelevant features can be present instead \cite{limitations_cond_generative_models, spatial_context_adv_robustness_obj_det_cvprw}.
Rather than specifying an exhaustive list of irrelevant features like rotation or color, as in data augmentation, removing the background eliminates many of them at once.

\PP{Mitigating Potential Issues}
\label{section:bg_rem_issues}
However, there are two potential issues.
First, trivial solutions can arise during training: the model can learn to localize objects in unmasked regions of the image, which is bad for generalization when we test with raw images because there will be many false positives in the background.
To avoid this, we break our training into three phases (Figure \ref{fig:defense_summary}).
The first phase is conventional fine-tuning to adapt the model to the domain.
The second is the originally proposed background-masked fine-tuning to learn relevant semantics and avoid spurious background signals.
The third mitigates trivial solutions: it implements the second phase but adds the original raw images into the background-masked training dataset for regularization, supervising that background regions should not be identified as objects purely because they are unmasked.
Training on this hybrid dataset thus aims to unlearn trivial solutions and suppress false positives, while preserving good existing solutions.

Second, background removal does not necessarily remove adversarial vulnerability.
Consider MNIST: images of digits on black background have no background, but can still produce models that are not robust.
The upshot is that most model vulnerability here is from background manipulations \cite{disentangle_adv_robustness_generalisation}, which pertains more to off-manifold digital attacks than physically-realizable attacks like our threat model.
To further increase robustness to the remaining on-manifold vulnerability, adversarial training on the manifold --- step 2 of OMAT --- has been shown to help \cite{disentangle_adv_robustness_generalisation}.
Next, we show how ARMOR does this without the instability of adversarial training.

\subsection{On-Manifold Fine-Tuning}
\label{section:on_manifold_finetuning}

\PP{Data Augmentation}
\label{section:on_manifold_data_aug}
To mitigate the low-data issue, we enrich the dataset with more training examples representing the invariances we wish to capture.
Many data augmentations (e.g., random noise, blur and object overlay) can surely be useful, but we aim to show that removing the background is already largely effective in removing irrelevant features, so we keep our augmentations modest.
We make 10 copies of the original image and randomly apply weather augmentations from \textcite{physical_aerial_adv_att} (increasing/decreasing light intensity, snow, rain, fog, autumn and no change) to the unmasked regions of images.
For parity, all evaluated defenses augment data with weather augmentations too unless specified.

\PP{Fine-Tuning with Random ON Patches}
\label{section:finetune_with_random_patch}
The second step of OMAT is adversarial training on the projected data, which adds noise optimized against the training objective to improve robustness.
However, optimizing noise in this min-max objective is unstable (Section \ref{section:off_and_on_manifold_attack_defense}), so we instead add uniform random noise.
Our approach is similar to randomized smoothing but used during training and not test-time --- we use random instead of optimized, adversarial noise to improve adversarial robustness.
Applied straightforwardly, this on-manifold noise would be added to unmasked regions of images.
However, given the threat model of adversarial patches that go onto objects rather than noise added to the entire image, we opt for patch insertion rather than noise addition, constraining these perturbations to be ON patches placed in the middle of the bounding box.
This way, we introduce an invariance where the model is \textit{not fooled by the presence} of a patch on the car, which is not naturally in the dataset, but is \textit{only fooled by an adversarial pattern} that is printed on the patch.
Hence, instead of adversarial training on the manifold, we perform our three-phase training which models training on the manifold (Section \ref{section:on_manifold_proj_bg_rem}) and add random noise in the form of random patches to all training samples.

\section{Experiments}
\label{section:adv_ml_exp}

To evaluate ARMOR as a defense, we ask four research questions: is ARMOR an effective \textbf{solution} for good clean \textit{and} adversarial performance [\textbf{RQ1}, Section \ref{section:main_results_digital_attack}]; \textbf{why} does it work [\textbf{RQ2}, Section \ref{section:ablations_advml}]; is it \textbf{generalizable} [\textbf{RQ3}, Section \ref{section:generalizability_tests}]; and is it \textbf{practically useful} [\textbf{RQ4}, Section \ref{section:adv_ML_realistic_evals}]?
To benchmark ARMOR, we use the set-up (e.g., attacks, models and dataset) primarily from \textcite{physical_aerial_adv_att}.

\subsection{Evaluation and Metrics}
\label{section:obj_det_eval_metrics}

Our main metric measures how successful the patch is in enabling the car to evade the object detector.
Hence, we measure the \textbf{average model confidence} (\textbf{Conf}, also known as objectness score) across all ground-truth objects, and also report the standard deviation.
Successful attacks decrease model confidence, while successful defenses mitigate that decrease.
To evaluate the trade-off between true positives, false positives and false negatives, we also use \textbf{average precision (AP)}, where higher AP corresponds to better separability between cars and everything else.
We report AP for our main results and ablations, where the false-positive trade-off is informative, and Conf elsewhere, since Conf directly measures attack success.

\subsection{Defending Digital Attacks}
\label{section:main_results_digital_attack}

\begin{figure*}[t]
    \centering
    \subfloat[Undefended.]{\includegraphics[width=0.148\linewidth]{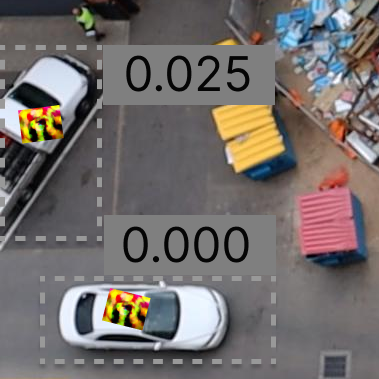}\label{fig:sidestreet_Undefended}}\hfil
    \subfloat[Fine-tuned.]{\includegraphics[width=0.148\linewidth]{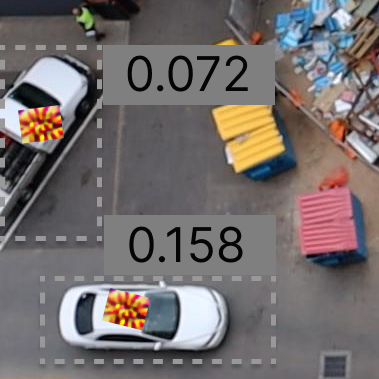}\label{fig:sidestreet_finetune}}\hfil
    \subfloat[RS \cite{randomized_smoothing_obj_det_neurips2020}.]{\includegraphics[width=0.148\linewidth]{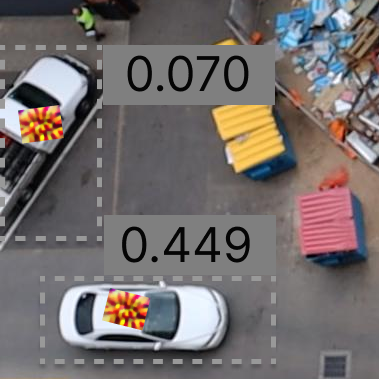}}\hfil
    \subfloat[PAD-S \cite{PAD_Jing_2024_CVPR}.]{\includegraphics[width=0.148\linewidth]{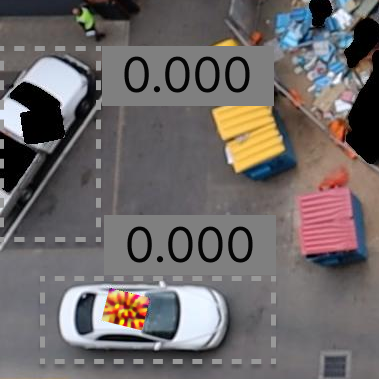}\label{fig:sidestreet_PAD}}\hfil
    \subfloat[SHIELD \cite{SHIELD_JPEG_Defense}.]{\includegraphics[width=0.148\linewidth]{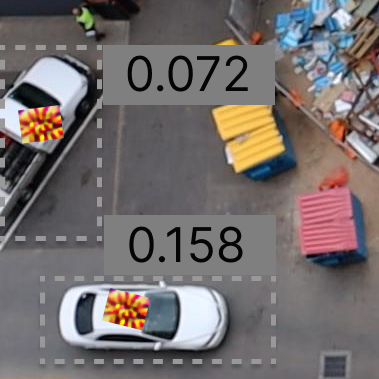}}\hfil
    \subfloat[ARMOR (Ours).]{\includegraphics[width=0.148\linewidth]{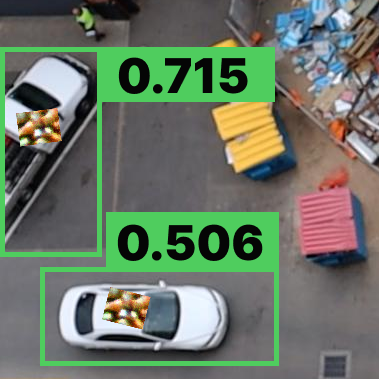}\label{fig:sidestreet_ARMOR_Defense_Aware}}
    \caption{\textbf{Main Results.} Sample visualizations of car detection by various defenses amidst defense-aware ON (ON-DA) attacks.
    Other defenses have predictions below the standard threshold of 0.50, resulting in no detections.
    Meanwhile, \textit{ARMOR mitigates the ON-DA attack and detects cars with high average confidence} of 0.62.}
    \label{fig:sidestreet_advON_method_comparison}
\end{figure*}

\begin{table*}[t]
\caption{\textbf{Main Results.} Conf and AP for each defense against no patches (clean), random and ON attack patches, comparing fine-tuning (FT), randomized smoothing (RS) \cite{randomized_smoothing_obj_det_neurips2020}, PAD \cite{PAD_Jing_2024_CVPR} without and with SAM (-S suffix), SHIELD \cite{SHIELD_JPEG_Defense} and ARMOR (ours); best ON-T and ON-DA results in bold.
    \textit{ARMOR maintains good Conf on all settings, achieving the best Conf against ON-T and ON-DA attacks by a large margin}.
    }
\centering
\resizebox{\textwidth}{!}{
\begin{tabular}{@{}l|cc|cc|cc|cc|cc|cc|cc@{}}
\toprule
\multirow{2}{*}{Patch$\backslash$Defense} & \multicolumn{2}{@{\hskip 3pt}c@{\hskip 3pt}|}{Undefended} & \multicolumn{2}{@{\hskip 3pt}c@{\hskip 3pt}|}{FT} & \multicolumn{2}{@{\hskip 3pt}c@{\hskip 3pt}|}{RS} & \multicolumn{2}{@{\hskip 3pt}c@{\hskip 3pt}|}{PAD} & \multicolumn{2}{@{\hskip 3pt}c@{\hskip 3pt}|}{PAD-S} & \multicolumn{2}{@{\hskip 3pt}c@{\hskip 3pt}|}{SHIELD} & \multicolumn{2}{@{\hskip 3pt}c@{}}{ARMOR (ours)} \\
\cmidrule(lr){2-15}
& Conf $\uparrow$ & AP $\uparrow$ & Conf $\uparrow$ & AP $\uparrow$ & Conf $\uparrow$ & AP $\uparrow$ & Conf $\uparrow$ & AP $\uparrow$ & Conf $\uparrow$ & AP $\uparrow$ & Conf $\uparrow$ & AP $\uparrow$ & Conf $\uparrow$ & AP $\uparrow$ \\
\midrule
Clean         & 0.83$\pm$0.16 & 0.89  & 0.96$\pm$0.04 & 1.00  & 0.91$\pm$0.10 & 1.00 & 0.94$\pm$0.04 & 1.00 & 0.96$\pm$0.04 & 1.00 & 0.96$\pm$0.05 & 1.00 & 0.92$\pm$0.07 & 1.00 \\
\midrule
Random ON     & 0.62$\pm$0.26  & 0.73  & 0.87$\pm$0.11  & 1.00  & 0.46$\pm$0.31 & 0.72  & 0.87$\pm$0.12 & 1.00 & 0.78$\pm$0.31 & 1.00 & 0.90$\pm$0.09 & 1.00 & 0.87$\pm$0.12 & 1.00 \\
ON-T Attack    & 0.14$\pm$0.19 & 0.19 & 0.36$\pm$0.27  & 0.99  & 0.31$\pm$0.32 & 0.54  & 0.48$\pm$0.38 & 1.00 & 0.48$\pm$0.37 & 1.00 & 0.42$\pm$0.33 & 0.94 & \textbf{0.73$\pm$0.16} & 1.00 \\
ON-DA Attack  & 0.14$\pm$0.19  & 0.19  & 0.14$\pm$0.14  & 0.95  & 0.30$\pm$0.29      & 0.57       & 0.14$\pm$0.26   & 1.00   & 0.17$\pm$0.28   & 1.00     & 0.14$\pm$0.14     & 0.95   & \textbf{0.62$\pm$0.20} & 0.97 \\
\bottomrule
\end{tabular}
}
\label{tab:digital_attacks_sidestreet}
\end{table*}

\PP{Set-Up}
There are 843 images from an aerial video feed of a sidestreet \cite{physical_aerial_adv_att}.
The first 780 images are used for training (fine-tuning) and the remaining 63 are reserved for testing.
Our ``baseline 0'' undefended model is a pre-trained YOLOv3 object detector \cite{yolov3} further pre-trained on COWC, an aerial imagery dataset of cars \cite{cowc_dataset}.
This model does not use the sidestreet training data, and is used as an initialization to other fine-tuned models.
During testing, we evaluate defenses on clean test images and with 3 types of patches: a random ON patch, controlling for each defense's vulnerability to the mere presence of a patch without adversarial patterns; the transfer ON (ON-T) attack patch, generated for the undefended model; and the defense-aware ON (ON-DA) attack patch, which stress-tests each defense against the strongest attack in our threat model (Section \ref{section:adv_att_threat_model}).
Sampling one \textit{diverse} method per defense category (Section \ref{section:adv_ML_def}), we compare ARMOR with (i) the undefended model, (ii) fine-tuning with weather augmentation, (iii) randomized smoothing \cite{randomized_smoothing_obj_det_neurips2020}, (iv) PAD \cite{PAD_Jing_2024_CVPR} with and without SAM, and (v) SHIELD \cite{SHIELD_JPEG_Defense}, omitting model-modification defenses due to their generalizability issues.
Defenses (iii)--(v) use the fine-tuned model as their base model, and we use the ON-DA patch from that fine-tuned model for their defense-aware evaluations.
We also experimented with patch localization methods Jedi and SAC, which had similar results to PAD, and with adversarial training via fine-tuning \cite{adv_training_obj_det, Shafahi_ICLR_2020_Adversarially_Robust_Transfer_Learning}, which was highly unstable and produced bad results on all settings; we exclude both for brevity.
We do not compare directly with OMAT because we focus on methods suitable for training data scarcity.

\PP{Main Results}
\label{section:main_results_sidestreet}
For \textbf{RQ1}, Table \ref{tab:digital_attacks_sidestreet} shows model confidence and AP across settings and defenses, and Figure \ref{fig:sidestreet_advON_method_comparison} shows sample visualizations.
The undefended model performs decently in clean settings (0.83 Conf, 0.89 AP), but is not robust to random or adversarial noise.\footnote{ON-T and ON-DA attacks are the same for the undefended model, based on how ON-T is defined.}
Its reduced AP suggests car confidence drops below other objects, increasing false positives --- expected due to lack of fine-tuning.
After fine-tuning, AP improves to $>$0.90 across most defenses, validating that fine-tuning removes those false positives, and the clean setting yields good performance across all defenses ($>$0.90 Conf and AP).
Random ON patches reduce model confidence for each defense, from 0.91--0.96 Conf to roughly 0.78--0.90 Conf, except for the undefended model and randomized smoothing.
Apart from ARMOR, the drop from clean to ON-T patches is even greater: 0.46--0.60, down to 0.31--0.48 Conf.
This is almost as large as the undefended model's 0.69 Conf drop from clean to ON-DA, showing that the ON attack is transferable from the undefended model to other defenses --- these defenses \textit{do not remove the vulnerability} they inherit from it.
Except for ARMOR, performance falls further under ON-DA patches to \textit{similar levels} as the undefended model at 0.14--0.30 Conf.
These defenses do not effectively address our physically-realizable threat model.

\PP{Baselines and Pre-processing Defenses}
Fine-tuning improves model confidence on both clean and random settings, corroborating our intuition that generalization and average-case robustness are correlated --- which supports ARMOR's choice to swap adversarial noise for random noise.
Randomized smoothing achieves poor model confidence of 0.46 and is the only defense with poor AP (0.72) for random ON patches, suggesting a lack of robustness even against average-case noise; it focuses on the $\ell_p$ threat model, which seems to break down for ON patches.
SHIELD performs well against random patches --- their high frequency can be alleviated with JPEG compression --- but cannot defend physically-realizable ON attack patches designed to be less pixelated, and thus lower frequency, obtaining a similarly low confidence to the fine-tuned model without SHIELD.
Although PAD/PAD-S is the SOTA for patch localization, it too has limited efficacy: Figure \ref{fig:sidestreet_PAD} shows PAD-S localizing only one of the two patches.
We posit that unrelated areas such as construction sites, and the smoothness of the attack patch, make patches look more ``normal'' and resist detection.
Even for the car with the masked patch, the detector still cannot detect it.
PAD is not enough as a defense.

\PP{ARMOR (Ours)}
ARMOR maintains good performance of over 0.90 Conf and AP in clean settings.
Although its 0.92 Conf is not as high as the fine-tuned baseline, it is far above the usual detection threshold of 0.50, so ARMOR remains as useful as other methods in practical settings.
Moreover, ARMOR has the smallest drop from clean to random ON patch, at 0.05 Conf, and the best confidence against ON-T by a large margin with 0.73 Conf.
The ON-DA patch still decreases ARMOR's performance, as expected when an adversary has more knowledge.
Nevertheless, against ON-DA attacks ARMOR is \textit{much better} than other defenses with 0.62 Conf compared to 0.14--0.30 Conf (Figure \ref{fig:sidestreet_ARMOR_Defense_Aware}), an improvement of 0.32.
Notably, ARMOR is more robust against the stronger ON-DA attack than other defenses are against the weaker ON-T attack (0.62 vs.\ 0.31--0.48 Conf).
ARMOR is an effective defense, maintaining good clean performance and much higher adversarial robustness than other defenses.

\subsection{Understanding ARMOR}
\label{section:ablations_advml}

\begin{table}[t]
    \caption{\textbf{Ablation.} Conf and AP for clean, ON-T and ON-DA settings.
    Row 1 is fine-tuned on the original, non-augmented dataset; later rows add weather augmentation (Wt), random patches (RP), or a fine-tuning phase with background-masked (BM) or hybrid (Hb) images.
    \textit{Random patches improve Conf under both ON-T and ON-DA}, and \textit{models trained without background} trade slight clean performance for \textit{much better ON-DA Conf}.
    Hybrid fine-tuning after BM fine-tuning mitigates false positives, shown by \ul{poor AP of BM fine-tuning alone} (underlined).
    }
    \centering
    \setlength{\tabcolsep}{4pt}
    \renewcommand{\arraystretch}{1.15}
    \resizebox{\linewidth}{!}{
    \begin{tabular}{@{}c c c c | c c c | c c c@{}}
        \toprule
        \multicolumn{4}{c|}{Ablated Settings} & \multicolumn{3}{c|}{Conf $\uparrow$} & \multicolumn{3}{c}{AP $\uparrow$}\\
         Wt & RP & BM & Hb & Clean & ON-T & ON-DA& Clean & ON-T & ON-DA \\
        \midrule
         &  &  &  & 0.89$\pm$0.10 & 0.18$\pm$0.23 & 0.11$\pm$0.19 & 0.98 & 0.39 & 0.27 \\
         \checkmark &  &  &  & 0.96$\pm$0.04 & 0.36$\pm$0.27 & 0.14$\pm$0.14 & 1.00 & 0.99 & 0.95 \\
         \checkmark & \checkmark &  &  & 0.96$\pm$0.05 & 0.52$\pm$0.34 & 0.25$\pm$0.21 & 1.00 & 0.92 & 0.92 \\
         \checkmark & \checkmark & \checkmark &  & 0.81$\pm$0.22 & 0.76$\pm$0.17 & 0.59$\pm$0.26 & \ul{0.64} & \ul{0.47} & \ul{0.30} \\
         \checkmark & \checkmark &  & \checkmark & 0.94$\pm$0.05 & 0.78$\pm$0.16 & 0.49$\pm$0.23 & 1.00 & 0.99 & 0.86 \\
         \checkmark & \checkmark & \checkmark & \checkmark & 0.92$\pm$0.07 & 0.73$\pm$0.16 & \textbf{0.62$\pm$0.20} & 1.00 & 1.00 & 0.97 \\
        \bottomrule
    \end{tabular}
    }
    \label{tab:ablation}
\end{table}

\subsubsection{Ablations}
To understand how ARMOR works [\textbf{RQ2}], we iteratively add ARMOR components to a vanilla fine-tuned model and evaluate each model.
We are most interested in the \textit{effectiveness} of background removal and random ON patches during training, as per our on-manifold adversarial training adaptation.
Results are in Table \ref{tab:ablation}.

\PP{Standard Fine-Tuning and Augmentations}
To contextualize background removal and random patches, we first understand what standard fine-tuning and weather augmentation can and cannot do.
Comparing the undefended model (Table \ref{tab:digital_attacks_sidestreet}) with the vanilla fine-tuned model (row 1, Table \ref{tab:ablation}), fine-tuning improves generalization, raising clean Conf from 0.83 to 0.89 with a slight ON-T improvement from 0.14 to 0.18.
Adding weather augmentation (row 2) continues the pattern --- clean Conf improves from 0.89 to 0.96 and ON-T from 0.18 to 0.36 --- as expected, since augmentation enriches the dataset and encourages generalization.
However, fine-tuning and augmentation \textit{do not} improve adversarial robustness: ON-DA performance remains poor at 0.11--0.14 Conf, similar to the undefended model.

\PP{Random Patch Augmentation}
Like regular augmentation, random ON patches during training (row 3) improve Conf in clean and ON-T settings.
Moreover, they significantly improve ON-DA Conf from 0.14 to 0.25, where weather augmentation alone only moved it from 0.11 to 0.14.
Like on-manifold adversarial perturbations in OMAT, random ON patches \textit{improve} both \textit{model generalization} and (on-manifold) \textit{adversarial robustness} --- and they do so while producing good clean performance, despite the \textit{distribution shift} that clean test images have no patches on them.
Hence, random ON patches are a \textit{practically effective adaptation} of on-manifold adversarial perturbation.

\PP{Background Removal}
Adding background-masked fine-tuning (BM FT: row 4) drastically improves ON-DA Conf from 0.25 to 0.59, showing that background removal \textit{improves} (on-manifold) \textit{adversarial robustness} and supporting our belief that it is a \textit{suitable and practical adaptation} of on-manifold projection in OMAT.
Replacing BM FT with hybrid fine-tuning (row 5) is not as effective --- ON-DA Conf decreases from 0.59 to 0.49 --- suggesting that adding raw images to BM FT is not as complete a model of on-manifold projection as pure BM FT.
However, AP after BM FT drops in all settings; investigating the predictions, we observe a lot of false positives, as predicted in Section \ref{section:bg_rem_issues}.
Nevertheless, hybrid fine-tuning \textit{after} BM FT (row 6) \textit{mitigates false positives} (high AP returns) \textit{while preserving adversarial robustness}, with ON-DA Conf rising slightly from 0.59 to 0.62.
Overall, ARMOR's three-phase training with raw, background-masked and hybrid images improves adversarial robustness while mitigating false positives.\looseness=-1

\subsubsection{Manifold Analysis}
\label{section:manifold_analysis}
OMAT assumes robustness arises from learning representations constrained to the data manifold (Section \ref{section:off_and_on_manifold_attack_defense}).
We verify this geometrically [\textbf{RQ2}] by comparing feature representations of the undefended model and ARMOR --- the two models that isolate ARMOR's training effect --- on three metrics.
We extract features from YOLOv3's penultimate layer (layer 81) for clean and patched test images, reduced to 30 dimensions via PCA.
For tangent space alignment, we use model-specific attack patches (ON-T for the undefended model, ON-DA for ARMOR) so that each model is probed by its own strongest perturbation.

\PP{Intrinsic Dimensionality}
A model focusing on task-relevant features rather than background noise should have fewer degrees of freedom in its representation.
Using the Maximum Likelihood Estimator \cite{maximum_likelihood_estimation}, ARMOR has \textit{lower intrinsic dimensionality} (1.64$\pm$0.54) than the undefended model (1.74$\pm$0.57).
This 5.7\% reduction suggests background removal \textit{constrains} the representation to \textit{car-relevant features}.

\PP{Local Linearity}
OMAT assumes the manifold is locally linear, allowing small perturbations to stay on it.
We measure this via reconstruction error: the variance unexplained by local PCA around each point.
ARMOR achieves \textit{lower} error (0.37$\pm$0.06) than the undefended model (0.41$\pm$0.14), indicating a more structured, locally linear manifold, and 57\% \textit{lower variance} (0.06 vs.\ 0.14), showing a more \textit{consistent} manifold structure across samples.

\PP{Tangent Space Alignment}
Finally, we measure the cosine similarity between perturbation directions and the local tangent space.
Both models show high mean alignment ($>$0.90), confirming that adversarial patches exploit on-manifold directions.
The difference lies in consistency: the undefended model is tightly aligned (0.96$\pm$0.04), so attacks reliably find and exploit the same manifold directions, whereas ARMOR shows 6$\times$ \textit{higher variance} (0.90$\pm$0.24) --- defense-aware attacks against ARMOR \textit{cannot} find consistent directions, and perturbations \textit{scatter} rather than converging on a single exploitable one.
Collectively, these metrics support our hypothesis that ARMOR's effectiveness stems from \textit{learning a more constrained, on-manifold representation}, explaining \textit{why} background removal and random patches improve robustness.

\subsection{Generalizability of ARMOR}
\label{section:generalizability_tests}

\begin{table}[t]
    \caption{\textbf{Generalizability.} Conf ($\uparrow$) across three shifts: random weather corruptions at test time, drone imagery of a carpark (541 images), and a much smaller detector (YOLOv11n) attacked with a patch trained on YOLOv3.
    \textit{ARMOR obtains the best Conf against ON-T and ON-DA attacks in every setting}.}
    \centering
    \resizebox{\linewidth}{!}{
    \begin{tabular}{@{}l l|c|c|c@{}}
    \toprule
    Setting & Patch & Undefended & FT & ARMOR (ours) \\
    \midrule
    \multirow{4}{*}{Weather}
    & Clean        & 0.63$\pm$0.39 & 0.82$\pm$0.28 & \textbf{0.85$\pm$0.22} \\
    & Random ON    & 0.47$\pm$0.34 & 0.65$\pm$0.33 & \textbf{0.73$\pm$0.29} \\
    & ON-T         & 0.23$\pm$0.25 & 0.35$\pm$0.32 & \textbf{0.61$\pm$0.27} \\
    & ON-DA        & 0.23$\pm$0.25 & 0.34$\pm$0.32 & \textbf{0.54$\pm$0.27} \\
    \midrule
    \multirow{4}{*}{Drone}
    & Clean        & 0.78$\pm$0.31 & \textbf{0.94$\pm$0.16} & 0.90$\pm$0.13 \\
    & Random ON    & 0.66$\pm$0.31 & \textbf{0.93$\pm$0.13} & 0.89$\pm$0.10 \\
    & ON-T         & 0.04$\pm$0.10 & 0.55$\pm$0.28 & \textbf{0.73$\pm$0.18} \\
    & ON-DA        & 0.04$\pm$0.10 & 0.43$\pm$0.26 & \textbf{0.71$\pm$0.19} \\
    \midrule
    \multirow{2}{*}{YOLOv11n}
    & Clean        & 0.12$\pm$0.26 & 0.66$\pm$0.11 & \textbf{0.68$\pm$0.21} \\
    & ON-T         & 0.13$\pm$0.16 & 0.55$\pm$0.15 & \textbf{0.68$\pm$0.13} \\
    \bottomrule
    \end{tabular}
    }
    \label{tab:generalizability}
\end{table}

We test ARMOR under three shifts [\textbf{RQ3}], reported together in Table \ref{tab:generalizability}.

\PP{Weather Corruptions in Test Data}
\label{section:weather_corruptions}
During deployment, test data may experience natural distribution shift due to image corruptions, such as weather (Section \ref{section:aerial_imagery_lit_review}), so we randomly sample weather corruptions per test image as we did for training (Section \ref{section:on_manifold_data_aug}).
Model confidence decreases across all models relative to our main results (Table \ref{tab:digital_attacks_sidestreet}); in particular, clean Conf under weather corruption degrades to each model's original performance on random ON patches \textit{without} corruption.
This correlation between robustness to global corruptions like weather and localized ones like random ON patches corroborates our insight that random ON patches during training improve feature robustness (Section \ref{section:finetune_with_random_patch}) [\textbf{RQ2}].
The attacks still land under corruption, reducing the undefended and fine-tuned model from 0.47 to 0.23 and 0.65 to 0.34--0.35 respectively, but ARMOR \textit{resists this drop the most} at 0.61 Conf against ON-T and 0.54 against ON-DA.

\PP{Different Setting: Drone Imagery}
\label{section:stationary_objects_carpark}
We next evaluate another form of aerial imagery: a moving drone surveying a densely populated carpark, where camera movement makes the background change constantly and introduces jitter.
Similarly, this dataset has only 541 images.
Again, the random ON patch decreases confidence for the undefended model with limited impact on the fine-tuned model and ARMOR.
The ON-T attack lowers the undefended and fine-tuned model to 0.04 and 0.55 Conf, and ON-DA further decreases the fine-tuned model to 0.43.
Meanwhile ARMOR stays well above the standard 0.50 detection threshold even under attack, at 0.90 Conf clean, 0.73 for ON-T and 0.71 for ON-DA.
Once again, \textit{ARMOR is robust while mitigating robust overfitting}.

\PP{Different Model}
\label{section:new_yolo_v11}
During deployment, defenders may prefer other detectors, and a common consideration is size: smaller models perform worse but are more energy-friendly, which is desirable in low-resource settings like drones.
Hence we evaluate YOLOv11 nano \cite{yolo11_ultralytics} (YOLOv11n), which takes 5\,MB of storage with 3M parameters, against 241\,MB and over 60M for YOLOv3.
As with YOLOv3, we pre-train on COWC, derive the undefended, fine-tuned and ARMOR models, and evaluate on the ON-T patch trained on YOLOv3 --- the realistic case where the attacker does not know the deployed model.
Clean performance drops for all YOLOv11n models: the smaller pre-trained model lacks expressivity and overfits to the pre-training data, though it remains useful for downstream fine-tuning, and both fine-tuned models stay above the usual 0.50 threshold.
Under the ON-T attack, the fine-tuned model drops about one standard deviation from 0.66 to 0.55 Conf, while ARMOR maintains \textit{high model confidence} of 0.68, demonstrating its \textit{generalizability to small models}.

\subsubsection{Stress-Test with Other Attacks}
\label{section:other_attacks}

We stress-test ARMOR against different defense-aware attacks by modifying the original attack [\textbf{RQ3}].
From the original ON-DA attack \cite{physical_aerial_adv_att} generated with BPDA \cite{adv_att_obfuscated_grad} (0.62 Conf), we change (i) the Adam optimizer to SGD (0.66), (ii) hybrid images to raw images (0.57), (iii) minimizing logits to minimizing the probability after sigmoid \cite{adv_att_aerial_agg_det} (0.57), (iv) minimizing the maximum detection confidence to minimizing the mean score \cite{adv_patch_att_mean_det} (0.52), and (v) patch initialization from the ON-T patch to random initialization (0.48).
The largest gain for the attacker comes from removing the ON-T warm start, which connotes that the warm start is not helpful and emphasizes that adversarial features for the undefended model \textit{cannot} be easily \textit{transferred} to ARMOR --- contrary to the transferability of ON-T patches for other defenses, both directly and as a warm start (Table \ref{tab:digital_attacks_sidestreet}).
Even this best defense-aware attack leaves ARMOR at 0.48 Conf, still better than every other defense under the \textit{original} ON-DA attack (0.14--0.30 Conf).
Overall, ARMOR attains 0.48--0.66 Conf under these attacks, \textit{more robust} than other defenses under ON-DA attacks and on par with them under the weaker ON-T attacks (0.14--0.48 Conf).

We also modify the threat model from patches on the car to patches around the car, known as ``OFF'' patches \cite{physical_aerial_adv_att}.
OFF patches can only be realistically implemented for stationary cars, limiting their deployability, and generating \textit{defense-aware} OFF patches is over 600 times more expensive than ON-DA patches due to their larger size and number, so we use the existing transfer OFF patch from \textcite{physical_aerial_adv_att}.
The OFF-T attack does not have much adversarial effect on ARMOR either, at 0.78 Conf compared to 0.73 Conf for ON-T, while other defenses obtain 0.21--0.66 Conf.
Hence, ARMOR is \textit{not overfitted} to the ON attack threat model, but also has potential to \textit{extend} to other \textit{physically-realistic} threat models.

\subsection{Evaluation on Realistic Physical Attacks}
\label{section:adv_ML_realistic_evals}

\begin{figure}[t]
    \centering
    \subfloat[Undefended.]{\includegraphics[width=0.28\linewidth]{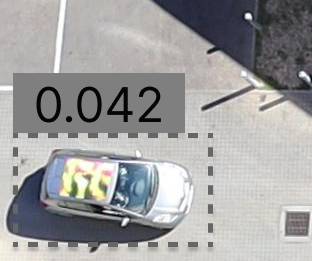}}\hfil
    \subfloat[Fine-tuned.]{\includegraphics[width=0.28\linewidth]{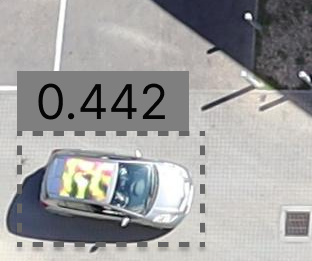}}\hfil
    \subfloat[ARMOR (Ours).]{\includegraphics[width=0.28\linewidth]{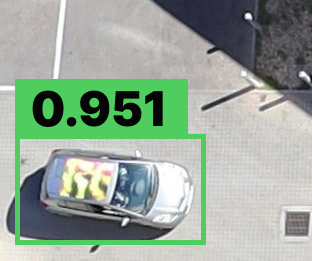}}
    \caption{\textbf{Realistic Physical ON-T Attack.}
    The undefended and fine-tuned model do not detect the car.
    In contrast, \textit{ARMOR detects cars with high average confidence} of 0.75.}
    \label{fig:sidestreet_phy_advON_method_comparison}
\end{figure}

\begin{table}[t]
    \caption{\textbf{Realistic Physical ON-T Attacks}. Conf ($\uparrow$) for printed patches affixed to car roofs, in the original sidestreet set-up and under motion blur.
    \textit{ARMOR obtains the best Conf while under realistic physical attack in both settings}.}
    \centering
    \resizebox{\linewidth}{!}{
    \begin{tabular}{@{}l l|c|c|c@{}}
    \toprule
    Setting & Patch & Undefended & FT & ARMOR (ours)\\
    \midrule
    \multirow{2}{*}{Sidestreet}
    & Clean & 0.83$\pm$0.16 & \textbf{0.96$\pm$0.04} & 0.92$\pm$0.07 \\
    & Physical ON-T & 0.01$\pm$0.09 & 0.38$\pm$0.37 & \textbf{0.75$\pm$0.23} \\
    \midrule
    \multirow{2}{*}{Motion blur}
    & Clean & 0.92$\pm$0.09 & \textbf{0.95$\pm$0.05} & 0.65$\pm$0.27 \\
    & Physical ON-T & 0.11$\pm$0.29 & 0.23$\pm$0.36 & \textbf{0.59$\pm$0.35} \\
    \bottomrule
    \end{tabular}
    }
    \label{tab:physical}
\end{table}

For \textbf{RQ4}, we evaluate settings closer to realistic deployment, constraining the attacker to the more realistic ON-T threat model.
Results are in Table \ref{tab:physical}.

\PP{Original Set-Up, but with Physical Attack}
\label{section:physical_attack_sidestreet}
We mimic a realistic attack-defense scenario in the original sidestreet setting, as in \textcite{physical_aerial_adv_att}: the attacker has access to training data and the pre-trained COWC model, then prints the patch and physically sticks it on the roof of the car.
Note the difference between the physical attack (Figures \ref{fig:adv_patches}b and \ref{fig:sidestreet_phy_advON_method_comparison}) and the digital one (Figure \ref{fig:sidestreet_Undefended}): the digital patch disregards the environment, being high resolution and acting as a source of light, whereas the physical patch is lower resolution and reflects environmental light.
Nonetheless, the physical ON-T patch is nearly as effective as the digital patch.
The undefended and fine-tuned models have low confidence (0.01 and 0.38), while ARMOR remains effective under \textit{realistic physical attacks}, maintaining \textit{high confidence} of 0.75.

\PP{Motion Blur}
\label{section:physical_attack_motion_blur}
During deployment, cars may have motion blur from fast movement, which may degrade attack efficacy by blurring the patch; we ask whether ARMOR offers more robustness than the baselines, or whether the attack simply stops working.
Motion blur is hard to generate with real cars due to safety concerns of driving fast in a small environment, so we use three palm-sized physical replicas driven by their pull-back motors.
Reusing the models fine-tuned on the original sidestreet dataset also simulates a scenario where the data collected for fine-tuning is insufficient.
In the clean setting the baselines have good zero-shot transfer, and although ARMOR's 0.65 Conf is not as high, it remains above the standard 0.50 threshold.
The undefended and fine-tuned models are \textit{vulnerable} to ON-T patches even \textit{with motion blur}, at 0.11 and 0.23 Conf, while ARMOR \textit{resists} the attack with 0.59 Conf.

\PP{Limitations and Extensions}
Under motion blur, ARMOR resists attacks but trails the baselines in the clean setting.
We suspect this is due to motion blur creating a distribution shift localized to object semantics: ARMOR forces the model to learn object-relevant features, which look different when training cars have no motion blur but test cars do.
ARMOR also assumes bounding box labels at fine-tuning time --- standard for detection, but it rules out fully unlabeled adaptation.
In future work, we hope to extend ARMOR to increase the robustness of learnt object-relevant features.

\section{Conclusion}

We address improving adversarial robustness of aerial object detectors under limited training data and propose ARMOR (Adversarial Robustness with Manifold-Oriented Training).
ARMOR modifies two data-intensive steps of on-manifold adversarial training.
First, it masks backgrounds using bounding boxes to approximate manifold projection.
Second, it inserts random patches on objects instead of optimizing adversarial noise.
Both steps draw only on labels the detection task already supplies, so the defense costs no extra annotation --- the binding constraint at a deployment site.
These changes encourage learning relevant, robust features while reducing data demands.
We evaluate ARMOR against realistic, physically-realizable patches and show it reduces robust overfitting and defends well, including against defense-aware attacks, and a manifold analysis of the learned features explains why.
Physical experiments confirm ARMOR is effective and practical.
We view ARMOR as a step toward making adversarial robustness practical.\looseness=-1

\bibliographystyle{IEEEtran}
\bibliography{bib_short, bib_bg_short, bib_CVPR_short}

\end{document}